\documentclass[review]{elsarticle}

\usepackage{lineno,hyperref}
\usepackage{subcaption}
\usepackage{adjustbox}
\usepackage{amsmath,verbatim,comment}
\modulolinenumbers[5]
\usepackage{graphicx}

\journal{Journal of Industrial Information Integration}

\begin{document}

\begin{frontmatter}

\title{Interaction models for remaining useful life estimation}

\author[1,2]{Dmitry Zhevnenko\texorpdfstring{\corref{cor1}}{}}
\ead{DmitryZhev@yandex.ru}
\address[1]{Artificial Intelligence Research Institute (AIRI), Moscow, Russia}

\address[2]{JCS MERI, Moscow, Russia}
\cortext[cor1]{Corresponding author}

\author[3]{Mikhail Kazantsev}
\ead{misha.kazantsev@gmail.com}
\address[3]{HSE University, Moscow, Russia}

\author[1,3,4]{Ilya Makarov}
\ead{iamakarov@misis.ru}
\address[4]{Big Data Research Center, National University of Science and Technology MISIS, Moscow, Russia}

\begin{abstract}
The paper deals with the problem of controlling the state of industrial devices according to the readings of their sensors. The current methods rely on one approach to feature extraction in which the prediction occurs. We proposed a technique to build a scalable model that combines multiple different feature extractor blocks. A new model based on sequential sensor space analysis achieves state-of-the-art results on the C-MAPSS benchmark for equipment remaining useful life estimation. The resulting model performance was validated including the prediction changes with scaling.
\end{abstract}


\end{frontmatter}

\newpage
\section{Introduction}

Modern Industry 5.0 trends determine high requirements for efficient monitoring and state control systems due to equipment complexity increase. Industrial data are characterized \cite{chen2016industrial, chen2020survey} by high variability from task and device, as well as the complexity of generating a sufficient number of anomalous states to build large expert models, similar to other domains \cite{devlin2018bert, ramesh2022hierarchical}. Thus, the development of the deep learning ``conduit-based maintenance" (CBM) models is related to domain-specific approaches for limiting data operation. The current methods described in the Related work section are limited to a few ideas with no possible expansion.  

Mostly, researchers focus on evaluation of the methods using C-MAPSS (NASA) dataset \cite{saxena2008damage}, which contains the aircraft engine state control problem \cite{lei2018machinery}. An accurate solution is not only to reduce the production costs associated with its operation but also to avoid failures situations during flight. 

We fixed the dataset preprocessing and compared current approaches and optimization methods using the base \cite{xu2013phm, sateesh2016deep, zheng2017long} simple models to evaluate the preprocessing importance. Having confirmed the preprocessing role, we describe the conditions sufficient for the performance comparison.

We proposed a technique to build a scalable model that combines different idea models as feature extractor blocks. We proposed a new model based on sequential sensor space analysis and scaled it according to the proposed methods achieving state-of-the-art (SOTA) results on the C-MAPSS dataset.

The main contributions of this work are as follows:

\begin{itemize}
    \item We proposed a novel feature extraction approach based on sequential modeling of sensor space, which achieved state-of-the-art results for the C-MAPSS dataset.
    
    \item We developed a scalable approach to a model extension using a combination of feature extractors on different operating principles and loss functions increasing the variety of extracting features.
    
    \item We proposed regularization techniques useful to prevent the overfitting specific to monotonic piece-wise RUL prediction problem.
\end{itemize}

The rest of the paper is organized as follows. The section ``Experiment Setting'' presents the dataset and its problem. The section ``Related work'' briefly gives the related research to the C-MAPSS dataset RUL problem. The section ``Proposed approach'' presents a detailed description of the proposed prognostic approach and used model structures. Section ``Experiments and Discussion'' includes a description of the learning process, experiments on the proposed models and their performance study. Finally, the ``Conclusion'' summarizes our research and includes possible feature work.

\section{Experiment Setting} 

The subsection includes the C-MAPSS dataset description, the physical meaning of containing data, the preprocessing of the feature and target spaces, and related metrics. 

\subsection{Dataset description}

The dataset \cite{saxena2008damage} includes the data generated by the ``Commercial Modular Aero-Propulsion System Simulation" (C-MAPSS). The program was used to simulate the behavior of turbofan engine sensors under various flight conditions until its failure. The flight conditions represent situations during ascent and descent to a standard flight altitude of about 10 km. 

The C-MAPSS dataset consists of four subsets of data, the details of which are presented in Table \ref{tab:cmapss_subsets}. Each subset contains readings from `21' sensors (Table \ref{tab:cmapss_sensors}) located in different parts of the degrading engine, which are controlled by three characteristics of the operational settings (height, speed and acceleration in terms of altitude, Mach number, and throttle resolver angle).

\begin{table}[h!]
    \centering
    \caption{Brief description of four C-MAPSS subdatasets.}
    \begin{tabular}{c c c c c }
        \hline \noalign{\vskip 0.1cm}
        Subdataset & FD001 & FD002 & FD003 & FD004 \\
        \noalign{\vskip 0.1cm} \hline \noalign{\vskip 0.2cm}
        Train trajectories & 100 & 260 & 100 & 249 \\
        Test trajectories & 100 & 259 & 100 & 248 \\
        Train samples & 20631 & 53759 & 24720 & 61249 \\
        Test samples & 13096 & 33991 & 16596 & 41214 \\
        Operating conditions & 1 & 6 & 1 & 6 \\
        Fault modes & 1 & 1 & 2 & 2 \\
        \noalign{\vskip 0.1cm} \hline
    \end{tabular}
    \label{tab:cmapss_subsets}
\end{table}

\begin{table}[h!]
    \centering
    \caption{C-MAPSS dataset engine sensors}
    \begin{adjustbox}{width=0.95\textwidth}
    \begin{tabular}{l l c}
        \hline \noalign{\vskip 0.1cm}
        Sensor \# & Description & Units \\
        \noalign{\vskip 0.1cm} \hline \noalign{\vskip 0.1cm}
        1 & Total temperature at the fan inlet & °R \\
        2 & Total temperature at the low-pressure compressor outlet & °R \\
        3 & Total temperature at the high-pressure compressor outlet & °R \\
        4 & Total temperature at the low-pressure turbine outlet & °R \\
        5 & Pressure at the fan inlet & psia \\
        6 & Total pressure in bypass-duct & psia \\
        7 & Total pressure at the high-pressure compressor outlet & psia \\
        8 & Physical fan speed & rpm \\
        9 & Physical core speed & rpm \\
        10 & Engine pressure ratio & — \\
        11 & Static pressure at the high-pressure compressor outlet (Ps30) & psia \\
        12 & Ratio of fuel flow to Ps30 & pps/psi \\
        13 & Corrected fan speed & rpm \\
        14 & Corrected core speed & rpm \\
        15 & Bypass ratio & — \\
        16 & Burner fuel-air ratio & — \\
        17 & Bleed enthalpy & — \\
        18 & Demanded fan speed & rpm \\
        19 & Demanded corrected fan speed & rpm \\
        20 & High-pressure turbine coolant bleed & lbm/s \\
        21 & Low-pressure turbine coolant bleed & lbm/s \\
        \noalign{\vskip 0.1cm} \hline
    \end{tabular}
    \end{adjustbox}
    \label{tab:cmapss_sensors}
\end{table}

The simulation was used to study the sensors during the engine's gradual deterioration up to the failure. Different noises were added to modeled engine sensor reading to approximate the numerical solution to the complicated real-world engine behavior \cite{saxena2008damage}. 

Similar to most of the work in the Related work section, we used normalize technique to the sensor readings. 

\subsection{Problem statement}

The common assumption to target space is that if the engine is functional, it is guaranteed to run to failure for a certain amount of time, which is determined by the constant value N:  

\begin{equation}
  RUL=\begin{cases}
    N, & \text{if $RUL \geq N$}\\
    RUL, & \text{if $RUL < N$}.
  \end{cases}
\end{equation}

$N$ is an heuristic parameter, and in most cases for the C-MAPSS dataset task $N = 125$ value is used due to previously performed empirical study. 
This piece-wise RUL target space, first proposed by Heimes in \cite{heimes2008recurrent}, is widely adopted in remaining useful life estimation tasks.  The sensor readings of a failing motor \ref{fig:sensor_reading} and the associated piece-wise RUL curve \ref{fig:piecewise-rul_fd1-u10} presented in Fig \ref{fig:sensor_RUL}.

Piece-wise RUL target function is at Fig. \ref{fig:piecewise-rul_fd1-u10}:

\begin{figure}[h!]
    \begin{subfigure}[b]{0.48\textwidth}
         \includegraphics[width=\textwidth]{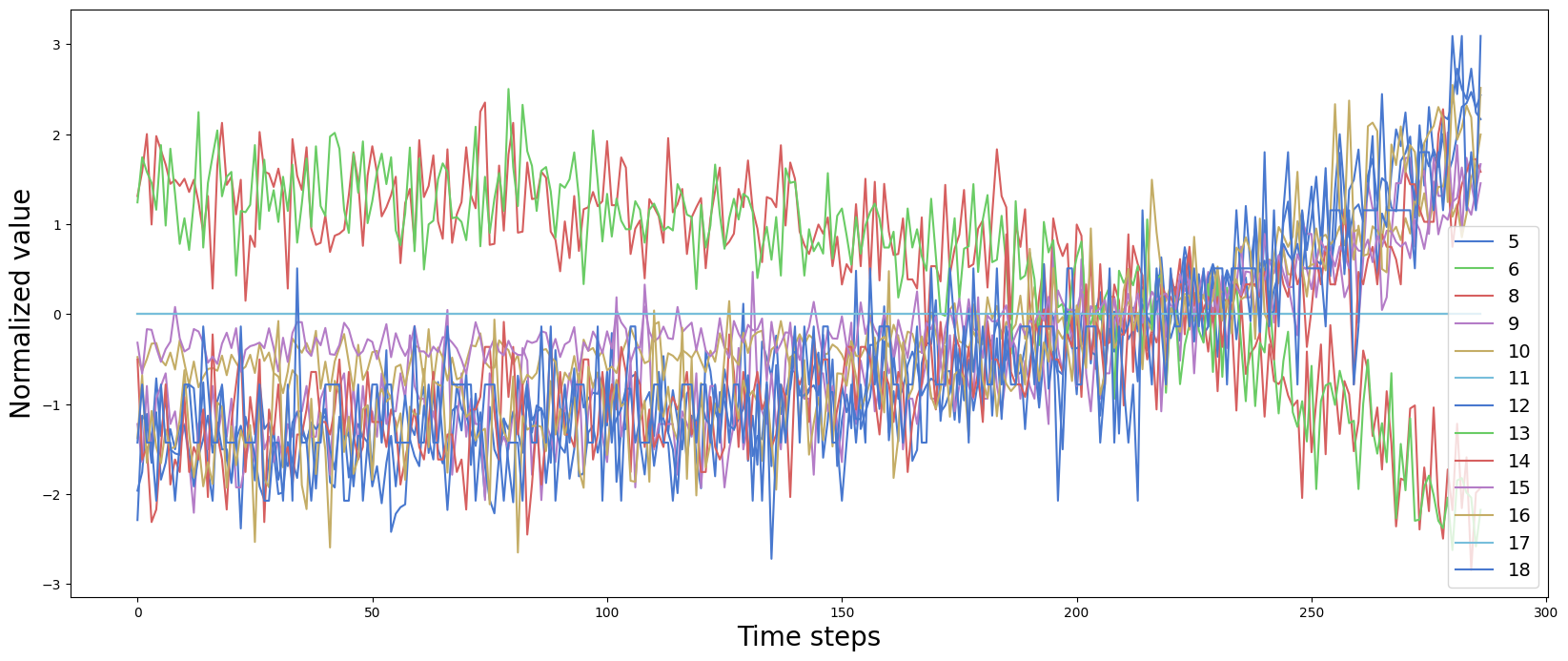}
         \caption[]{Sensor reading}
         \label{fig:sensor_reading}
     \end{subfigure}
     \hfill
     \begin{subfigure}[b]{0.48\textwidth}
         \includegraphics[width=\textwidth]{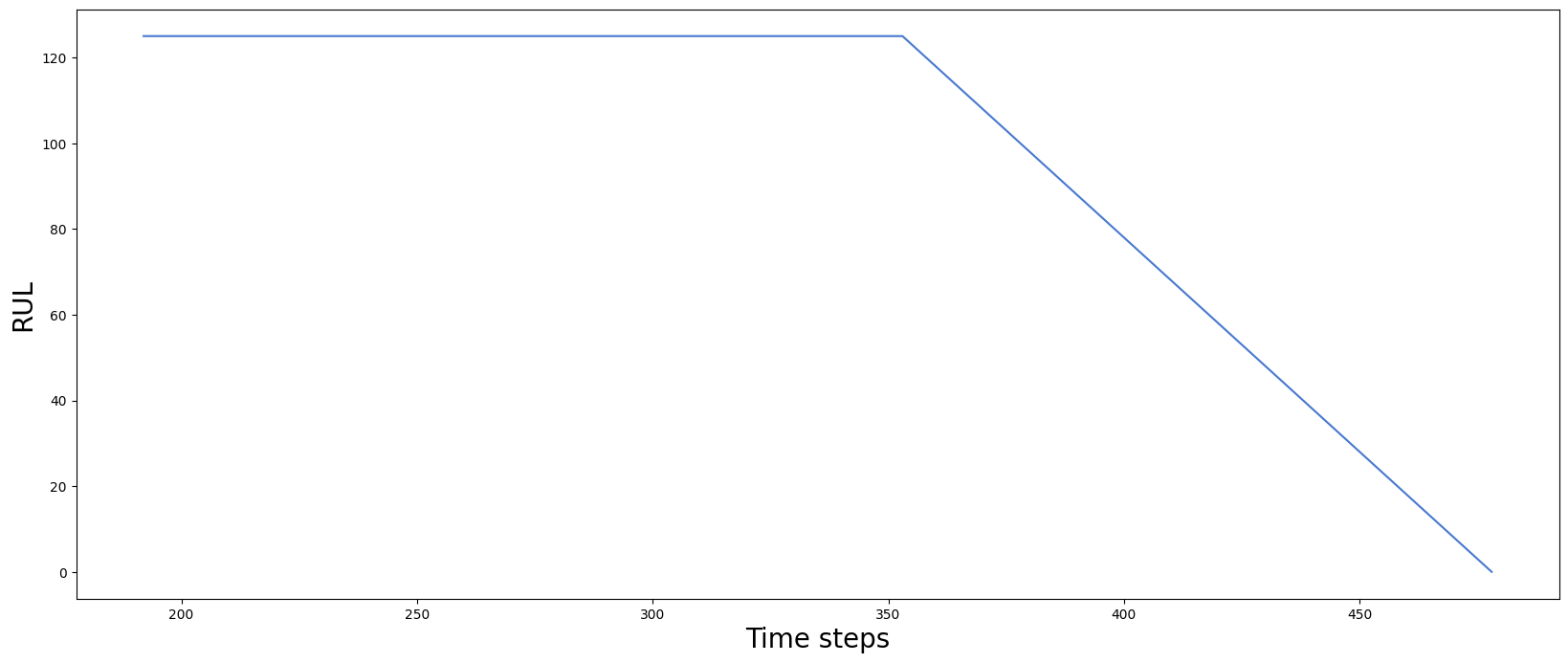}
         \caption[]{Piece-wise RUL}
         \label{fig:piecewise-rul_fd1-u10}
     \end{subfigure}
     \caption{Normalized sensor values and corresponding piece-wise RUL for the second engine of the fd001 subdataset}
     \label{fig:sensor_RUL}
\end{figure}

One of the problems limiting the model comparison ability is the use of different preprocessing. The severe unstable impact is caused by various preprocessing and postprocessing based on operation condition parameters (for example, \cite{pasa2019operating}). Thus, we considered only fd001 and fg003 sub-datasets with fixed operation conditions for proper model comparison. 

\subsection{Objective Quality Metrics}

The performance of the models on the C-MAPSS dataset was evaluated using RMSE and Score metric:

\begin{equation}
    RMSE = \sqrt{\frac{1}{n} \sum_{i=1}^{n}(y_i - \hat{y_i})^{2}}
\end{equation}

\begin{equation}
    \begin{gathered}
        \text{diff}(i) = \hat{y_i} - y_i\\
        score(i) = \begin{cases}
            e^{-\frac{\text{diff}(i)}{13}} - 1, & \text{if diff$(i) < 0$},\\
            e^{\frac{\text{diff}(i)}{10}} - 1, & \text{if diff$(i) \geq 0$}
        \end{cases}\\
        Score = \sum_{i=1}^{n}score(i)
    \end{gathered}
\end{equation}

where `${y_i}$' and $y_i$ are predicted and true value of RUL for engine `i'.

The asymmetric Score metric penalizes above the true value predictions due to maintenance regulations. The asymmetry is achieved via the coefficients in the denominator. These coefficients are empirically defined constants designed by the authors of the original dataset paper \cite{saxena2008damage} for the PHM2008 competition.

\section{Related Work}
There are two main ways for the C-MAPSS dataset RUL prediction problem: using auxiliary health index characteristics or a natural RUL curve.

\subsection{Auxiliary health characteristic}

The first method is domain-based auxiliary characteristics models adopted to map test samples to train-based latent space and to determine the relations of it to related train vectors to RUL prediction. One of the most common approaches is to construct a health index (HI) curve using deep learning and classical machine learning methods.

One of the main directions related to health index space modeling is the use of the similarity-based curve  matching technique, first presented in \cite{wang2008similarity}. The approach basis is the construction of multi-label time-series window mapping to the latent space. The latent space construction model is based on the training dataset, the prediction is performed by test samples mapping and distance to the closest train sample estimation.   

The article \cite{zhang2020aeroengines} uses sparse Bayesian learning (SBL) \cite{tipping2002analysis} to establish health indicators. In addition, there are methods of constructing an HI index based on classical ML clustering, proposed, for example, in \cite{javed2014robust}, where a new index is constructed using the sample distances to each cluster. 

One of the most important areas of HI generation is empirically constructed models, both in the form of classical equations \cite{malhotra2016multi} and deep learning approaches \cite{wang2008similarity, wei2021learning}. There are several Ways of constructing an HI index by the example of statistical approaches, e.g., in \cite{sentz2002combination} there are methods based on Dempster–Shafer evidence theory. Nevertheless, the construction of a plausible HI curve imposes serious requirements on the function or model to display. 

The HI curve can be generated using autoencoders of various \cite{yu2019remaining,duan2021bigru} principles for similarity-based curve matching techniques. In \cite{pillai2021two} as a feature extractor, a convolution-based autoencoder is used to construct a latent representation of the engine state that is maximally correlated with the function-based curve constructed from training sample sensor values. In the article, \cite{kim2021multitask} the authors proposed a comprehensive approach using a feature extractor built by applying multilayer CNN layers to a time-sequence sliding window. The main idea of the model is to construct a common feature space for both RUL prediction and device state classification. Some works contain elements of semi-supervised approaches based mainly on the use of different autoencoders \cite{berghout2020aircraft}.

\subsection{Direct RUL prediction}

The second method is a direct prediction of RUL or peace-wise RUL without using auxiliary features and reliance on domain-based knowledge based on DL models. The development of this direction is related to the possibility of using feature extractors and efficient optimizers and obtaining SOTA results on them. RUL prediction methods, in contrast to the health curves, estimate degradation processes, omitting local anomalies, for which other methods are developed and tested on specialized datasets \cite{da2020industrial, lomov2021fault}. In \cite{jayasinghe2019temporal} discusses an approach to generating partial training trajectories aimed at optimizing a linear RUL curve with a piece-wise continuous function that simulates the curve behavior in test datasets. In addition, the classical approach is to use some constant area of the RUL curve \cite{jayasinghe2019temporal, berghout2020aircraft} corresponding to the normal mode of the device, and the value of this area is usually chosen empirically. This heuristic is related to the fact that sensor readings in normal operation should not differ significantly from each other in the absence of noise - and thus model readings should not differ either. In addition to this assumption, the accuracy of the validation of the constructed model is also used to determine the value. Due to the various methods of estimation, most of the empirical estimates of RUL lie in the range of `115' to `160' switching cycles \cite{hong2020remaining, mo2021remaining, pillai2021two,li2018remaining, zheng2017long, ellefsen2019remaining, zhang2019remaining,ramasso2014review}.
The most popular approach to solving the RUL curve prediction problem is to use structures consisting of feature extractors and a regression head. The shape of a feature extractor varies widely, and for a regression head, MLP or gradient boosting are usually used. Early work in this area involves assessing the applicability of basic models like shallow networks and Multi-Layer Perceptrons (MLPs) \cite{gebraeel2004residual}. The authors of \cite{ zheng2017long} used LSTMs on three benchmark datasets and found them to work best compared to MLPs and CNNs. Wu et al. \cite{ wu2018remaining} compared LSTMs against vanilla RNNs and GRUs. In \cite{peng2020hybrid} feature extractor the authors present a series of heterogeneous heuristics on time series parameter extraction, and the use of LigthGBM to predict RUL and the associated curve, which the authors called a cumulative dynamic differential health indicator. Hong et al. in \cite{hong2020remaining} used stacked CNN, LSTM, and Bi-LSTM with residual connections to create a feature extractor, for which they considered approaches to increase explainability through dimensionality reduction and Shapley additive explanation techniques. In \cite{zhu2018estimation, zhao2017machine} the authors proposed the variants of feature extractors based on a multiscale convolutional neural network and a bidirectional recurrent block with control, respectively.

\subsection{Novel sota approaches}
In this section, several state-of-the-art models on various principles of operation are analyzed in terms of the proposed properties and ways to improve. 
As SOTA results, we would like to highlight the three following methods. 
The first principle \cite{costa2022variational, yu2021prediction} involves autoencoders and variational autoencoders for feature construction or creation of an aggregated health index device. Forecasting is usually achieved by using an lstm-type model, in which the principles of time transfer and information cleansing occur through stacked BiLSTM layers. As an example approach, we can consider \cite{costa2022variational}, a distinctive feature of which is the use of KL-div and MSE RUL loss to form a continuous ordered relative to RUL latent space, over which it is possible to perform state control by proximity similar to the similarity-based curve matching method discussed in the related work.
The second principle \cite{jin2022bi-lstm, xiang2022automatic,kara2022multi-scale} work includes feature construction approaches for further use in LSTM-based models. The use of different approaches to feature generation is so diverse that it includes not only the standard approaches. In particular, in \cite{jin2022bi-lstm} discusses the fusion of parallel models based on Bi-LSTM, with hand-crafted time-series features per model as input. The equipment state control is based on piece-wise RUL prediction. 

The third principle \cite{song2021distributed,narwariya2020graph,wang2021remaining} includes the addition of any form of the attention mechanism to the previously presented approaches. For example, in \cite{song2021distributed} attention mechanism was used both in the sensor allocation phase and in accounting for the influence of time series position.  The state of the equipment is supposed to be monitored based on RUL prediction. 

These papers, to a greater or lesser extent, share a common conception of cleansing and interaction of information within the input. However, each work relies on a few ideas with no possible expansion. We want to propose a new feature extraction approach and a method to use different models as extraction blocks to create a joint prediction. 

\section{Proposed Approach}

In this paper, we proposed a new approach focused on building blocks aimed at different features extracting and fusing the results to build high-level features. Different features extraction include selecting blocks of distinct nature and loss functions that ensure the distinction of selected output feature vectors before blending by the attention layer \cite{vaswani2017attention} into a common latent space.
Thus, the proposed approach includes different feature extraction blocks resulting in diverse latent space and attention layers to form high-level feature.

\begin{figure}[h!]
    \centering
    \begin{adjustbox}{width=\textwidth}
    \includegraphics{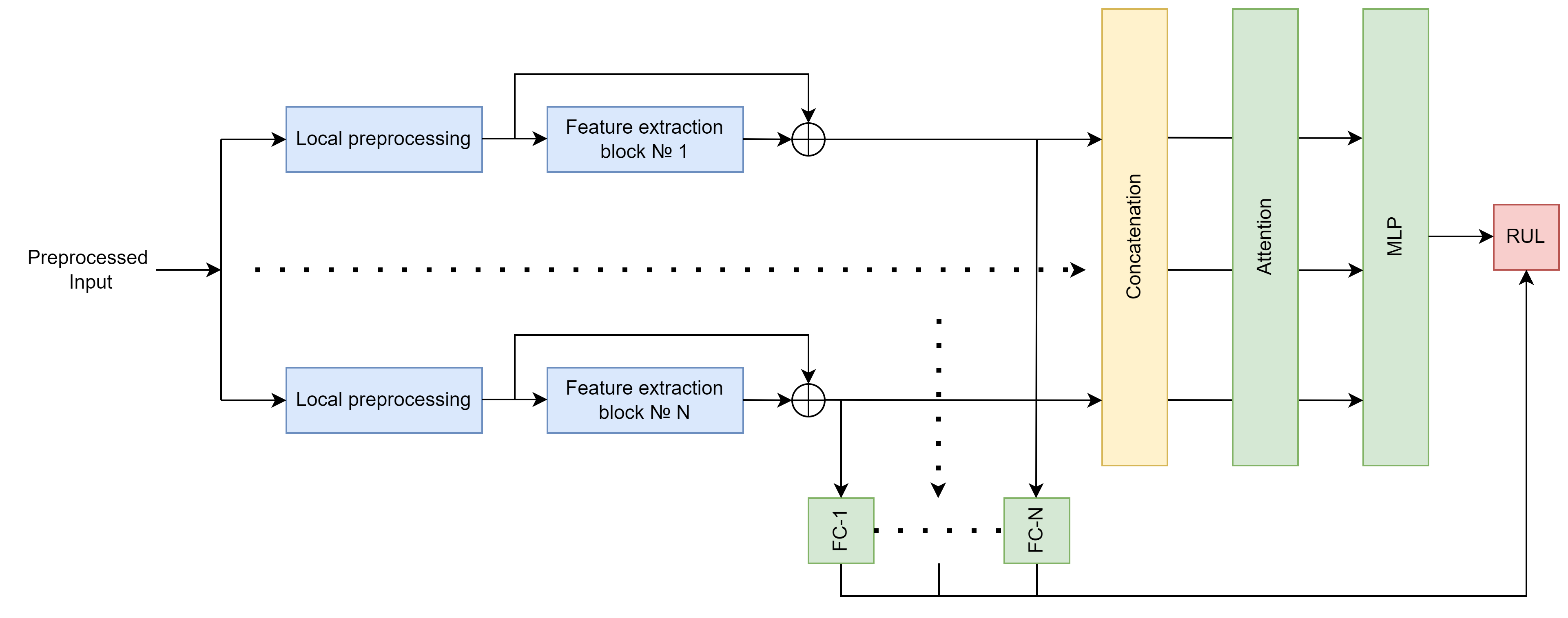}
    \end{adjustbox}
    \caption{Proposed method general structure}
    \label{fig:general_structure}
\end{figure}

The general structure of the model built within the proposed approach is shown at Fig. \ref{fig:general_structure}, and the neural network input goes through the following stages:

\begin{itemize}
    \item Dataset-related preprocessing

    \item Preprocessed time-series is passed to the input of ``Local preprocessing'' in form of attention to specific features or time points, generalization or refinement of the sequence, and focusing on the time or feature domain.
    
    \item The resulting sequence is passed to the feature extraction block with the skipped connection, allowing to maintain the gradient distribution regardless of the block structure.
   
    \item Latent space vectors at the output of the feature extraction blocks are normalized and accounted for in loss, increasing the distinction between them.
 
    \item Each normalized vector passes through additional fully connected layers for RUL prediction so that at the expense of the general error to form a selection of this level important features.
    
    \item The resulting different vectors are concatenated and passed through an attention layer to compensate for unavoidable errors in the form of noise associated with the requirement for vector difference.
  
    \item The resulting latent representation passes through the regression head in the form of a multilayer perceptron for RUL prediction.
\end{itemize}

The described approach resulted in the following step-by-step created structures. First, we studied the effect of the chosen data preprocessing methods on LSTM and CNN based models. Second, we evaluated the effect of input data transposition as local preprocessing. Finally, we built three consecutive models, increasing the number of feature extractor blocks according to the proposed framework. 

\subsection{Baseline setup}

We evaluated how much the use of the proposed preprocessing can affect the simulation results. To investigate the effect of dataset based preprocessing heuristics, we use the well-known models proposed for the CMAPSS dataset problem as described in Related Work section. For that, we have reimplemented models based on Long Short-Term Memory (LSTM) and Convolution Neural Network (CNN), which have been state-of-the-art models for solving the C-MAPSS dataset task shortly after it emerged.

\begin{figure}[h!]
    \centering
    \begin{adjustbox}{width=\textwidth}
    \includegraphics{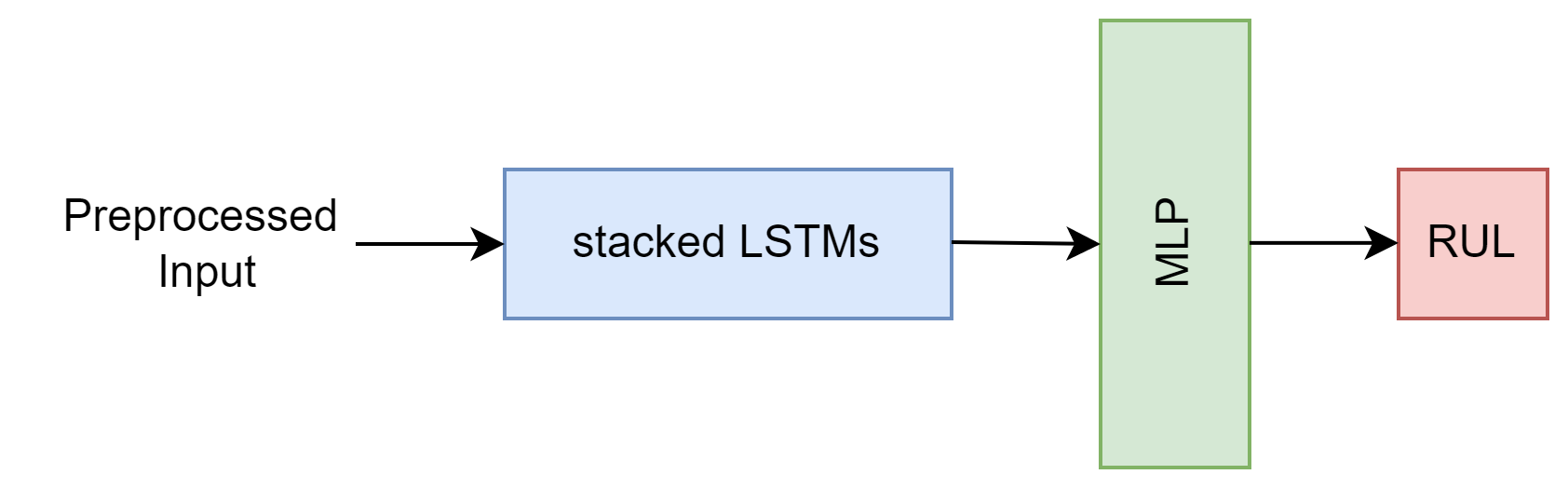}
    \end{adjustbox}
    \caption{LSTM-based general structure}
    \label{fig:lstm-arch}
\end{figure}

Figure \ref{fig:lstm-arch} demonstrates the general structure of LSTM-based approach. ``Stacked LSTMs'' bock consists of three stacked LSTM layers with the parameters: hidden state and input features is `21', dropout is `0.5'. ``MLP block'' consists of two fully-connected layers $(441, 126)$ and $(126, 1)$ with ReLU nonlinearity and 0.5 dropout.

\begin{figure}[h!]
    \centering
    \begin{adjustbox}{width=0.9\textwidth}
    \includegraphics{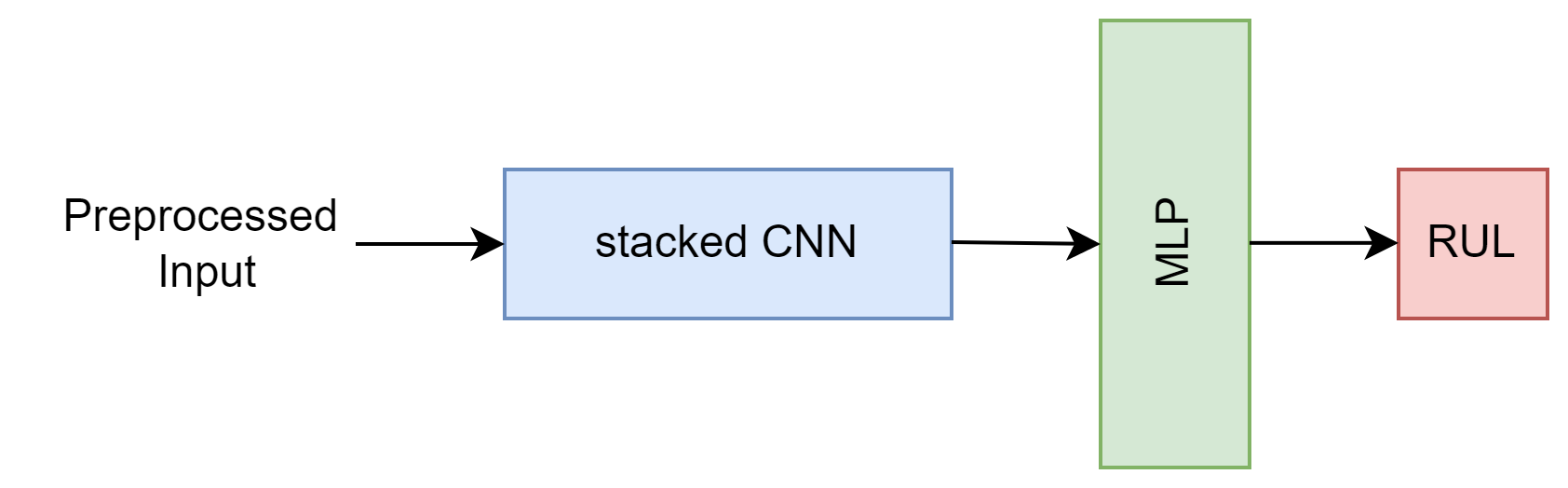}
    \end{adjustbox}
    \caption{CNN-based general structure}
    \label{fig:cnn-arch}
\end{figure}

In turn, Convolutional Neural Network is an algorithm, usually employed for image processing. But due to their structure and ability to capture spatial and temporal dependencies, they are widely used for time series forecasting as well. Figure \ref{fig:cnn-arch} illustrates the general structure of the implemented CNN-based approach. ``Stacked CNN'' includes two consecutive blocks of 1d Convolution (kernel sizes are `7' and `3', out channels are `5' and `10', stride sizes are `1' and `0'), ReLU non-linearity, max pooling (kernel sizes are `3' and `3', stride sizes are `2' and `1'), and flatten procedure after them. ``MLP block'' consists of three fully-connected layers $(30, 60)$, $(60, 120)$ and $(120, 1)$ with ReLU nonlinearity and 0.5 dropout. 

\subsection{Feature approach}

We investigated one feature extractor block of the proposed approach. The effectiveness of the "local preprocessing" block was tested for the first time through a swap of the time and feature domains. A study for window sizes one and three showed that domain swap leads to a 20\% change in score metrics while maintaining comparable RMSEs, so we investigated the effect by adding that transform to the extractor block-like structure. 

\begin{figure}[h!]
    \centering
    \begin{adjustbox}{width=\textwidth}
    \includegraphics{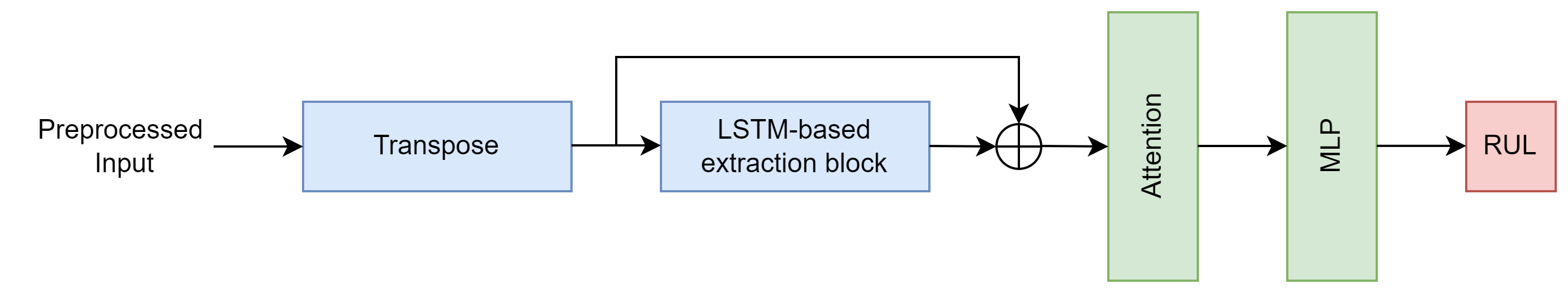}
    \end{adjustbox}
    \caption{Time-feature model structure}
    \label{fig:Time_feature_block}
\end{figure}

The structure of the time-feature model (TFM) for impact evaluation studied is shown in \ref{fig:Time_feature_block}. Block “Transpose” is time and feature domains swap, “LSTM-based extraction block” is a block of 3 stacked LSTMs with implementation of skipped connection (feature size is equal to hidden size) with parameters dropout `0.5', feature size is equal to window size `32' and `40' for fd001 and fd003 subdatasets, self-attention implementation corresponds to \cite{vaswani2017attention}. ``MLP block'' consists of two fully-connected layers $(672, 256)$, $(256, 1)$ with SiLU nonlinearity \cite{elfwing2018sigmoid} and `0.5' dropout. 

\subsection{Time-feature interaction}

Next, we tested how effective the proposed Interaction approach is in increasing the prediction accuracy compared to using the exact single model. The model presented in the previous paragraph was used as one of the feature extractor blocks because of its high efficiency.The first step is to test the use of double TFM (DTFM), relying on the separation of different features through the use of an modified cosine loss function: 

\begin{equation}
    \cos{(z_i,z_j)} = \frac{z_i z_j}{\max{(\|z_i\|_2 \|z_j\|_2, \epsilon)}}
\end{equation}

\begin{equation}
    mCosineLoss =\sum_{(i,j), i\neq{j}}\max{(\cos{(z_i,z_j)}, 0)}
\end{equation}

where $\epsilon = 1e^{-7}$ is to avoid zero in the denominator, $z_k$ is the kth extractor block output vector. 

The use of Huber losses reduces the effect of local variance in predicting a nonnatural monotonic RUL curve shape:

\begin{equation}
    \begin{gathered}
        HuberLoss(i) = \begin{cases}
            \frac{1}{2}(y_i - \hat{y_i})^{2}, & \text{if $|(y_i - \hat{y_i})| < \beta$},\\
            
            \beta(|(y_i - \hat{y_i})|- \frac{1}{2}\beta), & \text{otherwise}
        \end{cases}\\
    \end{gathered}
\end{equation} 

where hyperparameter $\beta = 1$.

Thus, the overall loss function consists of three parts, representing the loss on RUL predictions based on concatenated vectors, losses on RUL predictions based on feature extractor blocks latent vectors, and differences between that latent vectors:

\begin{equation}
L = HuberLoss_{model} + \lambda \cdot \sum_{k}HuberLoss_{k} + \sigma \cdot mCosineLoss
\end{equation}

where the empirical parameters lambda and sigma depend on the number of feature extractor blocks and the batch size, `k' is the feature extractor block index and are equal `0.3' and `1' correspondingly, `model' is index of final `MLP' output (Fig \ref{fig:general_structure}). 

\begin{figure}[h!]
    \centering
    \begin{adjustbox}{width=\textwidth}
    \includegraphics{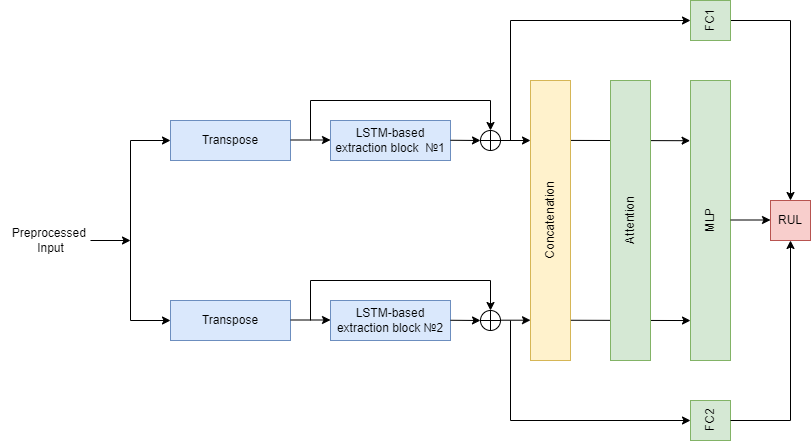}
    \end{adjustbox}
    \caption{Double time-feature model structure}
    \label{fig:DTime_feature_block}
\end{figure}

The structure of the model for impact evaluation is shown in Fig. \ref{fig:DTime_feature_block}. The model blocks are described above, except: ``Concatenation'' is concatenate feature extractor block outputs, ``FC1'' and ``FC2'' are fully-connected layers $(672, 1)$ with SiLU nonlinearity \cite{elfwing2018sigmoid} and `0.5' dropout, ``MLP block'' consists of two fully-connected layers $(1344, 256)$, $(256, 1)$ with SiLU nonlinearity \cite{elfwing2018sigmoid} and `0.5' dropout. The dimensionality of the hidden vector is much larger than the window size and applying blocks of the same nature further seems redundant.

\begin{figure}[h!]
    \centering
    \begin{adjustbox}{width=\textwidth}
    \includegraphics{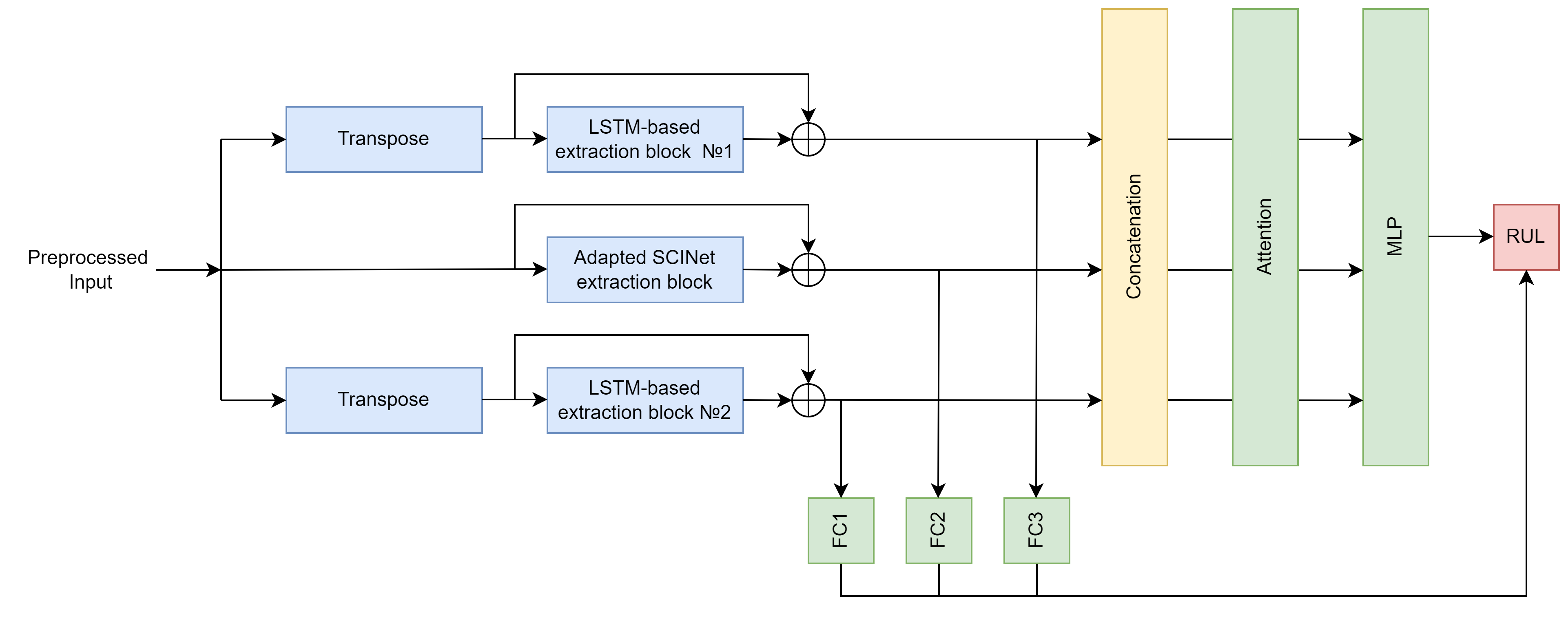}
    \end{adjustbox}
    \caption{Time-feature interaction model structure}
    \label{fig:TFIM_model}
\end{figure}

\begin{figure}[h!]
    \centering
    \begin{adjustbox}{width=0.8\textwidth}
    \includegraphics{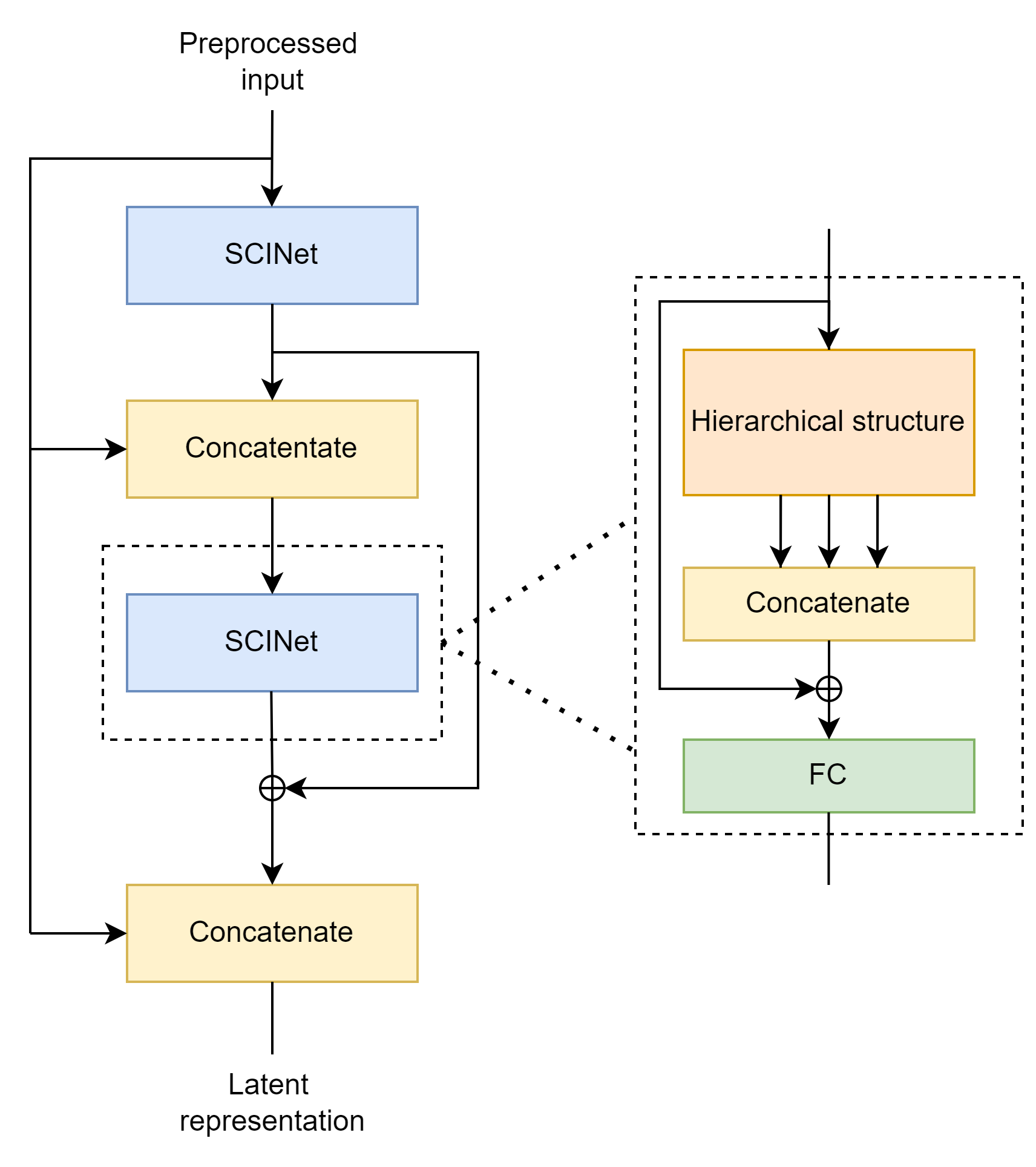}
    \end{adjustbox}
    \caption{Adapted SCINet extraction block structure}
    \label{fig:scinet}
\end{figure}

To further improve the metric, we adapted state-of-the-art stacked SCINet \cite{liu2021time} as an additional future extractor (Fig. \ref{fig:TFIM_model}) to build the time-feature interaction model (TFIM). The reimplemented structure includes an adaptation of the multilayer hierarchical model (Fig. \ref{fig:scinet}), where "Hierarchical structure" represents convolution-based affine interactions, detailed in \cite{liu2021time}. The main parameters of the built block are as follows: hidden size is `16', three levels in the hierarchical structure, kernel size is `3', and dropout is `0.65'. Other parameters staing constant except ``MLP'' first layer change to $(2016, 256)$.

Adaptation of the proposed block to the task of RUL prediction consists in optimization of the latent layer for RUL prediction between stacked blocks at the expense of additional skipped connections through the adapted SciNet block, and using the approach for processing the output vector in accordance with the proposed method.

\section{Experiments and Discussion}

\subsection{Training procedure}

In this section, we represented the learning process, the validation, the features of overfitting, and the hyperparameter choice.

Due to the described dataset properties, one of the important tasks is regularization. In addition to mentioned class imbalance, there are anomalous functioning states in the data, realized by random outliers in the engine working regime. Thus, the importance of regularization increases. In this paper, regularization was performed by four main effects: reducing the class imbalance during dataset formation by considering only sixth window with $RUL=125$, adding white noise to the training sample, using high dropout values, and averaging several RUL predictions to reduce variance. 

Thus, summarizing the proposed regularization methods for the problem and the traditional approaches described in Related work, we fix the following conditions: 

\begin{itemize}
    \item Selecting a sliding window of sizes `32' and `40' for fd001 and fd003, the insufficient test sample (engines number `1' and `52' for fd001 and fd003 accordingly) expanded to the left.
    \item AdamW \cite{loshchilov2017decoupled} as the optimizer.
    \item Huber loss to all RUL prediction blocks.
    \item Data noise augmentation (Normal distribution sampling, $\sigma = 0.04$)   .
    \item Averaging over five values to reduce variance in prediction.
    \item Averaging over five runs to reduce variance in training.
\end{itemize}

Basic models were trained with a sliding window size of `32' and batches of size `128'. The adaptive triangular cyclic learning rate scheduler \cite{smith2017cyclical} with [$1e^{-4}$; $5e^{-4}$] interval. 

The optimal learning configuration for training Interaction models include `210' batch size and adaptive triangular cyclic learning rate updating rule in [$2e^{-4}$, $9e^{-5}$]. 

The validation includes the selection of `37' (fd001) and `45' (fd003) consecutive switching points (to average over `5' sliding windows of sizes `32' and `40'), which are excluded from the random location of each training sample engine and used for five-fold cross-validation with division by engines.

This approach reduces the training datasets by about `5\%' due to `37' and `45' time points per engine exclusion and leads to a scoring decrease due to the high diversity of trajectories in feature space. The choice of other validation techniques leads to even more significant deterioration. However, due to the proposed regularization method, there was a plateau with minimal metric values and a number of epochs to reach for every model. Thus, the proposed models were trained on the full dataset and the epoch number obtained on validation. To obtain the final result, the predictions on the test dataset averaged over five epochs after reaching the required epoch. On average, the training procedure lasted over `120' epochs, reaching a plateau around `45' or `70' depending on the model. 

Once the minimum was reached, the overfitting began. One interesting effect that can be associated with overfitting is a decrease in prediction accuracy as the final characteristic (the RUL prediction curve for all engine readings) approaches the reference peace-wise form. This effect can be attributed to the ``close" trajectories \cite{saxena2008damage} in the sensor reading space due to the presence of multiple noises added during the dataset data simulation. 

The non-monotonicity of the characteristic guarantees a more accurate hit in the mean, but automatically enhances the variance of the result. The variance impact of the model was reduced by averaging the last five predictions with the correction for the successive reduction of the RUL parameter.

\subsection{Domain preprocessing impact}

This section includes the performance of the discussed preprocessing approach to well-known base models (Table \ref{tab:old-sota-metrics}). 

\begin{table}[ht!]
    \centering
    \caption{State-of-the-art approaches of previous years with a performance, comparable to baselines performance from this thesis}
    \begin{adjustbox}{width=\textwidth}
    \begin{tabular}{l p{0.3cm} cc p{0.3cm} cc}
        \hline \noalign{\vskip 0.2cm}
        & & \multicolumn{2}{l}{FD001} & & \multicolumn{2}{l}{FD003} \\
        \noalign{\vskip 0.05cm}
        \cline{3-4} \cline{6-7}
        \noalign{\vskip 0.1cm}
        & & RMSE & Score & & RMSE & Score\\
        \noalign{\vskip 0.2cm} \hline \noalign{\vskip 0.2cm}
        MODBNE (2017) \cite{zhang2017multiobjective} & & 17.96 & 640 & & 19.41 & 683\\ \noalign{\vskip 0.1cm}
        Deep LSTM (2017) \cite{zheng2017long} & & 16.14 & 338 & & 16.18 & 852 \\ \noalign{\vskip 0.1cm}
        BiLSTM-ED (2019) \cite{yu2019remaining} & & 14.74 & \textbf{273} & & 17.48 & 574 \\ \noalign{\vskip 0.1cm}
        Attention-based DL Approach (2021) \cite{chen2021machine} & & 14.53 & 322 & & \textemdash & \textemdash \\ \noalign{\vskip 0.1cm}
        CNN (ours) & & \textbf{14.38} & 332 & & 15.12 & 495 \\ \noalign{\vskip 0.1cm}
        LSTM-baseline (ours) & & 14.88 & 400 & & \textbf{14.75} & \textbf{382} \\ \noalign{\vskip 0.1cm}
        \noalign{\vskip 0.2cm} \hline
    \end{tabular}
    \end{adjustbox}
    \label{tab:old-sota-metrics}
\end{table}

Table \ref{tab:old-sota-metrics} shows the comparison results of the constructed models with the models that used other preprocessing and learning methods. These works were considered state-of-the-art, but the use of described approaches on base models has surpassed their results. The base models outperformed the referenced models in terms of RMSE and nonsymmetrical score constructed according to state control requirements. 
Thus, this study demonstrated the importance of actual general learning and preprocessing methods for a fair comparison.   

\subsection{Time-feature interaction experiments}

Using the approach described above, we trained the models presented in the Time-feature interaction section. The training results in comparison to state-of-the-art results are in Table \ref{tab:sota-metrics}. 

\begin{figure}[]
     \begin{subfigure}[b]{0.49\textwidth}
         \includegraphics[width=\textwidth]{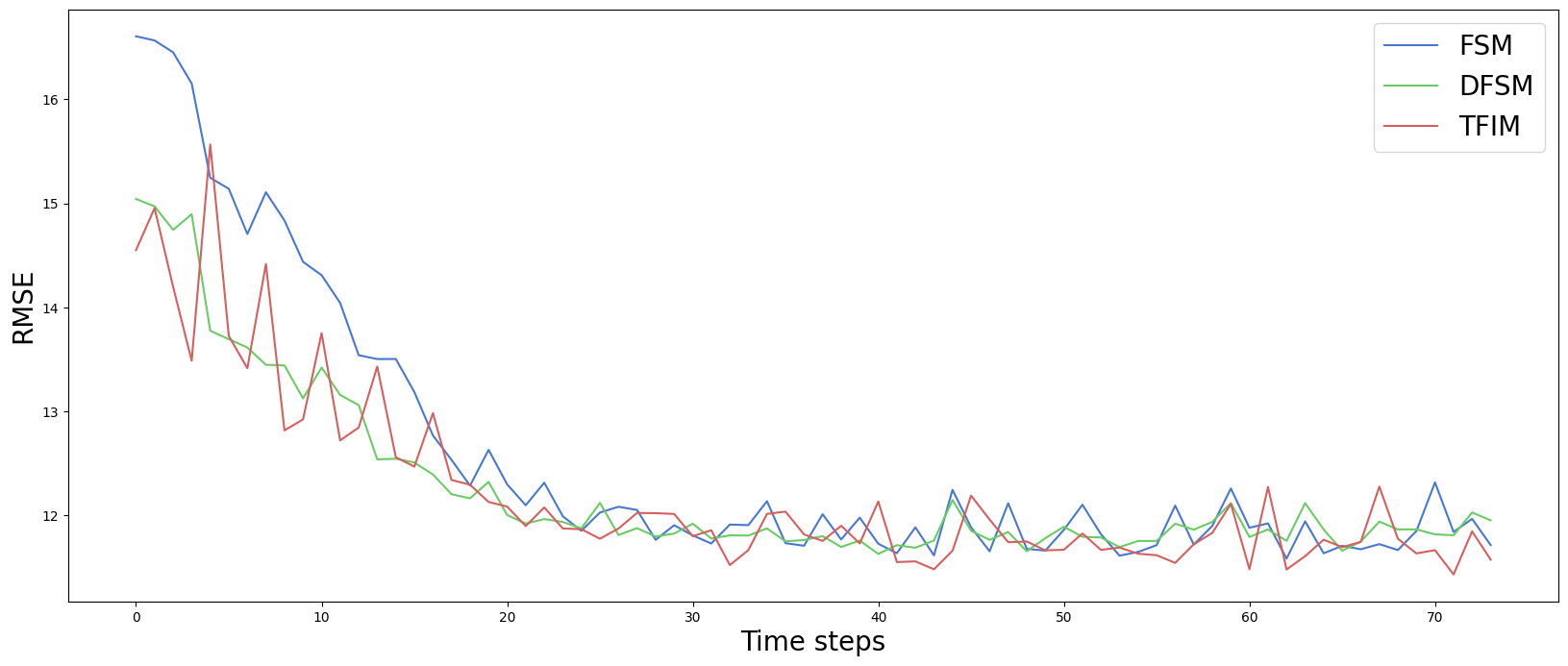}
         \caption[]{Learning curves}
         \label{fig:learning_curves}
     \end{subfigure}
     \hfill
     \begin{subfigure}[b]{0.49\textwidth}
         \includegraphics[width=\textwidth]{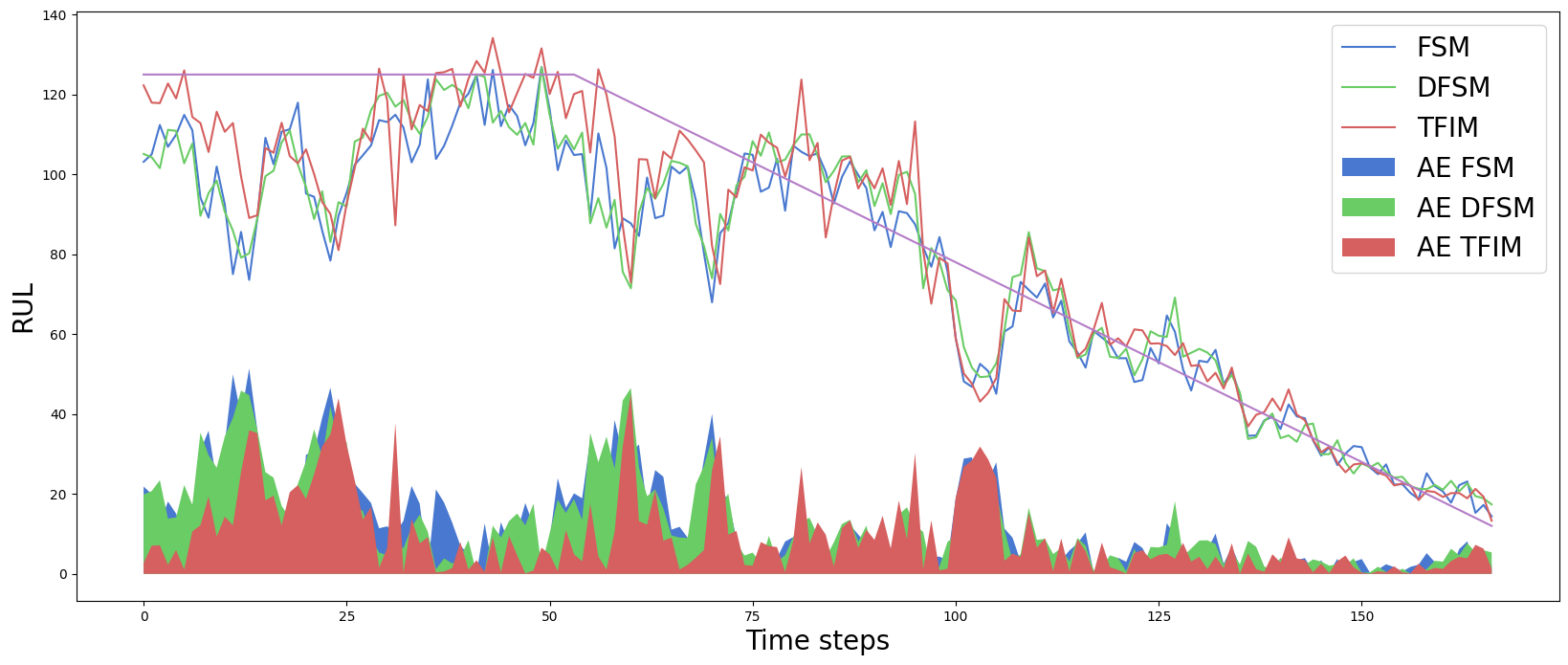}
         \caption[]{35th test engine prediction}
         \label{fig:test_engines}
     \end{subfigure}
     \caption{Learning curve and prediction result for 35th engine of dataset fd001}
     \label{fig:learning}
\end{figure}

The performance of the models on the standard for demonstration engines and model training are shown in Fig. \ref{fig:test_engines} and \ref{fig:learning_curves}. The 35th engine is used to show the prediction results of each of the three models considered, AE is the absolute error between the model prediction and the true RUL value at that point.

Each of the proposed models showed state-of-the-art results. However, when moving from a model with one feature extractor block to a model with three feature extractor blocks, the RMSE changed by less than `2\%', although the inference time, measured by `300' runs on a dummy example using the GPU (NVIDIA A100)  warm-up technique, changes from `0.01' to `0.75' seconds. 

The standard deviation in the table is composed of the variation of the metric value at the five plateau points, over which we average the performance, and the variation between five consecutive runs to measure the stability of the model.

\begin{table}[t]
    \centering
    \caption{Recent C-MAPSS dataset state-of-the-art approaches}
    \begin{adjustbox}{width=\textwidth}
    \begin{tabular}{l p{0.3cm} cc p{0.3cm} cc}
        \hline \noalign{\vskip 0.2cm}
        & & \multicolumn{2}{l}{\hskip 0.58cm FD001} & & \multicolumn{2}{l}{\hskip 0.5cm FD003} \\
        \noalign{\vskip 0.05cm}
        \cline{3-4} \cline{6-7}
        \noalign{\vskip 0.1cm}
        & & RMSE & Score & & RMSE & Score \\
        \noalign{\vskip 0.2cm} \hline \noalign{\vskip 0.2cm}
        GNMR (2020) \cite{sota1} & & 12.14 & 212 & & 13.23 & 370 \\ \noalign{\vskip 0.1cm}
        BLS-TFCN (2021) \cite{sota2} & & 12.08 & 243 & & 11.43 & 244 \\ \noalign{\vskip 0.1cm}
        TCNN-Transformer (2021) \cite{sota4} & & 12.31 & 252 & & 12.32 & 296 \\ \noalign{\vskip 0.1cm}
        RVE (2022) \cite{sota6} & & 13.42 & 323 & & 12.51 & 256 \\ \noalign{\vskip 0.1cm}
        MCA-BGRU (2022) \cite{sota8} & & 12.44 & 211 & & \textemdash & \textemdash \\ \noalign{\vskip 0.1cm}
        DA-TCN (2021) \cite{sota9} & & 11.78 $\pm$ 0.29 & 229 $\pm$ 8 & & 11.56 $\pm$ 0.61 & 257 $\pm$ 58 \\ \noalign{\vskip 0.1cm}
        ADL-DNN (2022) \cite{sota7} & & 13.05 $\pm$ 0.16 & 238 $\pm$ 5 & & 12.59 $\pm$ 0.25 & 281 $\pm$ 5 \\ \noalign{\vskip 0.1cm}
        Feature space model (ours) & & 11.73 $\pm$ 0.09 & 215 $\pm$ 5 & & 11.64 $\pm$ 0.03 & 191 $\pm$ 9\\ \noalign{\vskip 0.1cm}
        Double feature space model (ours) & & 11.65 $\pm$ 0.07 & 210 $\pm$ 6 & & 11.58 $\pm$ 0.03 & 204 $\pm$ 6 \\ \noalign{\vskip 0.1cm}
        Time-feature interaction model (ours) & & \textbf{11.59 $\pm$ 0.08} & \textbf{208 $\pm$ 3} & & \textbf{10.9 $\pm$ 0.06} & \textbf{187 $\pm$ 8} \\ \noalign{\vskip 0.1cm}
        \noalign{\vskip 0.2cm} \hline
    \end{tabular}
    \end{adjustbox}
    \label{tab:sota-metrics}
\end{table}

\begin{figure}[h!]
    \centering
    \begin{adjustbox}{width=\textwidth}
    \includegraphics{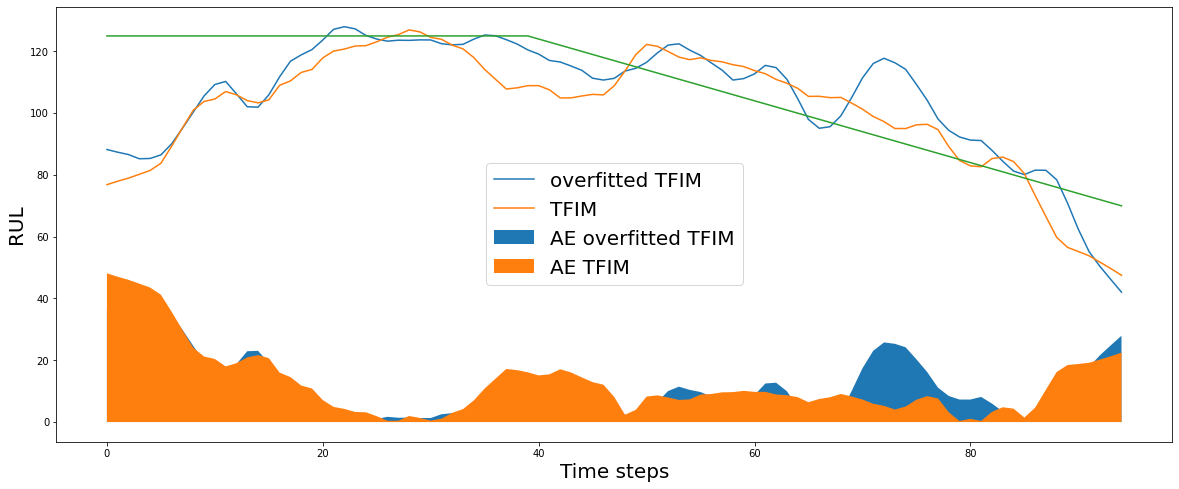}
    \end{adjustbox}
    \caption{Smoothed prediction of RUL engine `2' by trained and overfitted models}
    \label{fig:Overfitting}
\end{figure}

It is important to note that models with non-monotonic noisy predictions obtained the best results. A typical example of the overfitting that occurs after exceeding the plateau is shown in Fig. \ref{fig:Overfitting}. Increasing the accuracy of modeling the curve shape near-constant values significantly reduces the prediction accuracy in the degradation region, which leads to a decrease in the metric on the test dataset. We attribute this process to the unnatural shape of the piece-wise RUL curve, which forces the model to learn how to convert noisy trajectories into a monotonically decreasing curve, creating a noticeable loss. 

\begin{figure}[h!]
    \centering
    \begin{adjustbox}{width=\textwidth}
    \includegraphics{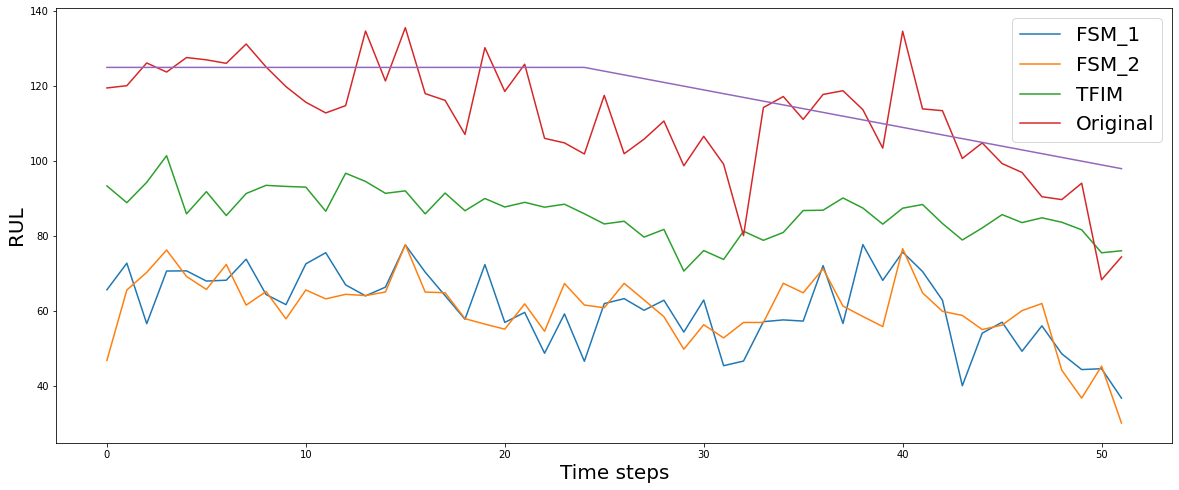}
    \end{adjustbox}
    \caption{TFIM 10th engine prediction with zeroed outputs of related feature extractor block}
    \label{fig:Inference}
\end{figure}

To check if all the features are used in the prediction, we perform a zeroing of the part of the latent representation corresponding to each feature extractor block \ref{fig:Inference}. The presented result demonstrates that the interaction of both FSM and adapted SciNet blocks are qualitatively similar, which can be interpreted as the extraction of close features. However, due to the use of loss on the difference of vectors, locally both predictions are different, which suggests the extraction of additional information. The combined information usage from all three blocks allows us to obtain a qualitatively and quantitatively close prediction.

\begin{figure}[h!]
    \centering
    \begin{adjustbox}{width=\textwidth}
    \includegraphics{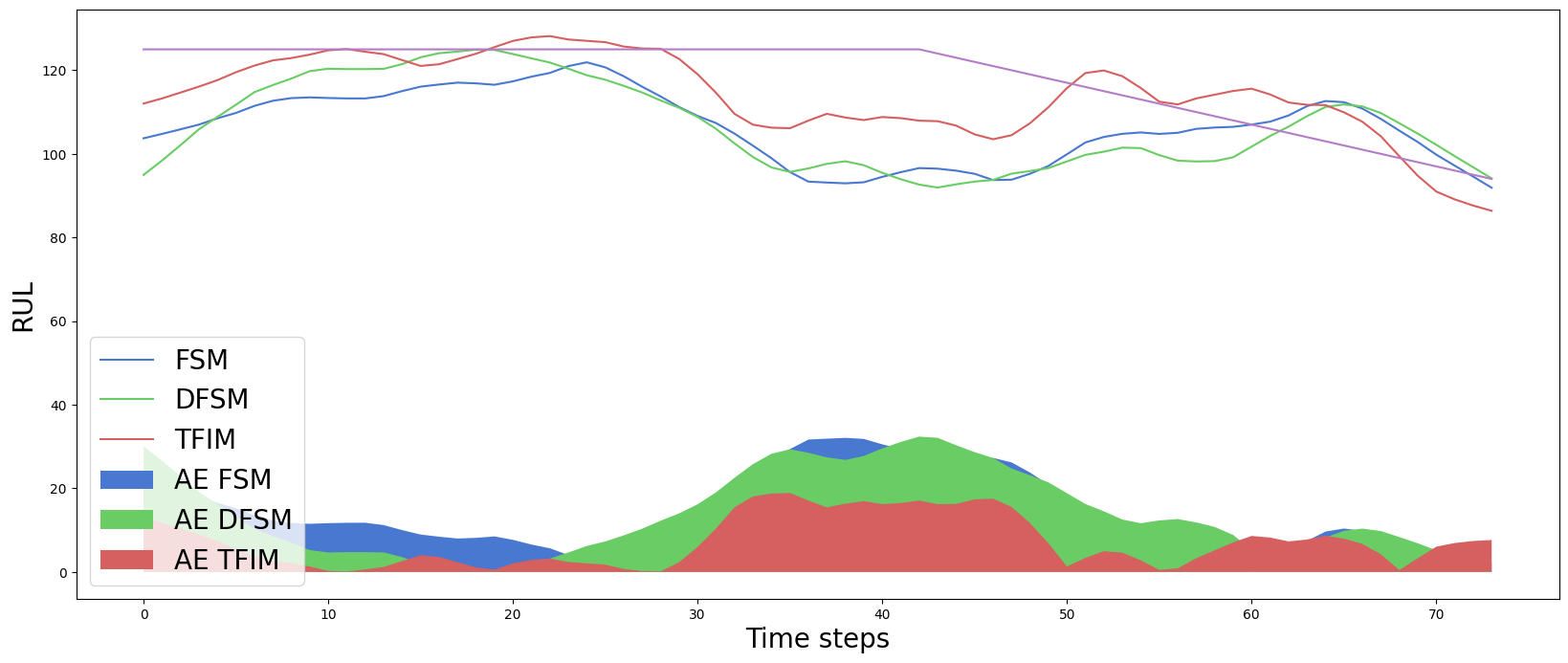}
    \end{adjustbox}
    \caption{Smoothed RUL prediction for the difficult to simulation 6th engine}
    \label{fig:worst}
\end{figure}

During the comparison study of three models, we found that the use of an additional feature extractor block primarily affects the generalizability. To confirm this effect, we selected test engine number `6' from the fd001 subdataset, on which the models show one of the worst results, and over-smoothed it using the ``Locally Weighted Scatterplot Smoothing" method with the frac parameter `0.11', essentially retaining only the overall trend (Fig. \ref{fig:worst}). The engine was chosen as an example of prediction changes corresponding to the improved quality of the models. The figure shows that TFIM supports state predictions with $RUL = 125$ in the working state and contains a clear degradation trend in the transition region, unlike other models. A similar pattern is observed for most of the test engines, indicating that further improvements in model quality are possible.

\section{Conclusion}

In this study, we proposed the "Interaction" approach to the state control problem on the example of the C-MAPSS dataset. The method includes building scalable models based on blocks of distinct nature and loss functions that ensure the distinction of selected output feature vectors. 
Using multiple feature extractor blocks for feature extraction leads to scalability and interpretability due to the simplification of block individual inference research. In the framework, we have explicitly proposed a dataset preprocessing, the effectiveness of which was measured relative to basic models. Also, overfitting for the C-MAPSS dataset models was described as a model adaptation to the monotonic piece-wise RUL curve prediction.

We proposed a new Feature space model (FSM) that demonstrated state-of-the-art results for the C-MAPSS task in selected preprocessing and augmentations approaches. The resulting model was used to form blocks of general Interaction structure. The approach to different latent space vector formation building allowed the doubled FSM (DFSM) significantly improve the metric. Adding an adapted SCINet model to the structure demonstrates the possibility of extracting additional information at the expense of a block of a different nature. The constructed time-feature interaction model (TFIM) not only improves the metrics but also leads to an increase in the generalizability of the model.  
Further work on developing the approach could be to find rules for scaling the system in-depth with the development of regularization methods that compensate for the drawbacks of the piece-wise constant RUL form.

\bibliography{bibliography}

\end{document}